# CATER: Leveraging LLM to Pioneer a Multidimensional, Reference-Independent Paradigm in Translation Quality Evaluation

**Kurando IIDA**, **Kenjiro MIMURA**

ErudAite. Inc.

Tokyo, Japan

## Abstract:

This paper introduces the Comprehensive AI-assisted Translation Edit Ratio (CATER), a novel and fully prompt-driven framework for evaluating machine translation (MT) quality. Leveraging large language models (LLMs) via a carefully designed prompt-based protocol, CATER expands beyond traditional reference-bound metrics, offering a multidimensional, reference-independent evaluation that addresses linguistic accuracy, semantic fidelity, contextual coherence, stylistic appropriateness, and information completeness. CATER's unique advantage lies in its immediate implementability: by providing the source and target texts along with a standardized prompt, an LLM can rapidly identify errors, quantify edit effort, and produce category-level and overall scores. This approach eliminates the need for pre-computed references or domain-specific resources, enabling instant adaptation to diverse languages, genres, and user priorities through adjustable weights and prompt modifications. CATER's LLM-enabled strategy supports more nuanced assessments, capturing phenomena such as subtle omissions, hallucinations, and discourse-level shifts that increasingly challenge contemporary MT systems. By uniting the conceptual rigor of frameworks like MQM and DQF with the scalability and flexibility of LLM-based evaluation, CATER emerges as a valuable tool for researchers, developers, and professional translators worldwide. The framework and example prompts are openly available, encouraging community-driven refinement and further empirical validation.

# 1. Introduction

Machine translation (MT) evaluation has long depended on reference-based, string-matching metrics such as BLEU (Papineni et al., 2002) or edit-distance variants like TER (Snover et al., 2006). While these metrics accelerated early MT research, their reliance on static references and inability to capture contextual, stylistic, or informational subtleties have become increasingly problematic. Neural-based metrics, including COMET (Rei et al., 2020), have improved correlations with human judgments but remain limited in scope. They often focus narrowly on semantic fidelity, leaving critical dimensions—such as discourse coherence, stylistic appropriateness, and domain-specific completeness—insufficiently addressed.

The advent of large language models (LLMs) opens a new era of translation evaluation. LLMs, including GPT-4 and other advanced transformers, possess broad multilingual capabilities and can understand textual nuances well beyond surface-level correspondences. They offer the unprecedented opportunity to perform dynamic, context-aware evaluations without rigid reliance on a single reference translation. Yet to fully harness LLMs for evaluation, we need a principled framework that integrates multiple quality dimensions and translates them into actionable, quantifiable scores.

This paper presents the Comprehensive AI-assisted Translation Edit Ratio (CATER), a next-generation MT evaluation framework that seamlessly integrates with LLMs through prompt-based instructions. CATER moves beyond single-metric evaluations by targeting five key dimensions—Linguistic Accuracy (LA), Semantic Accuracy (SA), Contextual Fit (CF), Stylistic Appropriateness (STA), and Information Completeness (IC)—each grounded in established quality frameworks like MQM and DQF. Crucially, CATER's approach is operationalized via an LLM: the user provides a standardized prompt, source text, and target text, and the LLM identifies required edits, calculates category-specific Edit Ratios (ER%), and outputs dimension-specific and overall scores.

This direct LLM integration yields several immediate benefits:

Reference independence: CATER dispenses with single canonical references, allowing for the evaluation of diverse or creative MT outputs.
Multilingual and domain adaptability: By simply adjusting prompt instructions or weights, evaluators can tailor CATER to any language pair, domain, or stylistic requirement.
Scalability and ease of implementation: No custom models or extensive training data are required. Any environment supporting LLM-based inference can host a CATER evaluation session.
Contextual and stylistic sensitivity: Through the LLM's inherent contextual understanding, CATER more effectively captures subtleties like discourse-level coherence, cultural tone, or factual completeness.

While future research will systematically benchmark CATER against established metrics, this paper lays the conceptual and practical foundation for a new paradigm in MT quality assessment. We argue that CATER's readiness for immediate deployment—via LLM prompts—can reshape how translation quality is measured and managed, benefiting researchers, industry practitioners, and global translation communities alike.

## 2. Background and Related Work
### 2.1 Traditional Metrics
- **BLEU (Papineni et al., 2002):** BLEU, or Bilingual Evaluation Understudy, was one of the first widely adopted metrics for evaluating machine translation. It measures the n-gram precision between the machine translation output and one or more reference translations. While rapid and widely used, its simplistic overlap measures fail to account for paraphrastic variation or discourse-level factors.
- **TER (Snover et al., 2006):** Translation Edit Rate (TER) measures the number of edits needed to change a machine-translated sentence into the reference sentence. Edits can include insertions, deletions, substitutions, and shifts. TER introduces edit-distance calculations to estimate post-editing effort. TER's conceptual foundation informs CATER, but TER's dependence on a single reference and lack of multidimensional evaluation remain limiting.
- **METEOR (Banerjee & Lavie, 2005):** Metric for Evaluation of Translation with Explicit ORdering (METEOR) was designed to address some of the weaknesses in BLEU. It uses a combination of exact word matching, stemming, synonym matching, and paraphrase matching to align and score the machine translation against a reference.

### 2.2 Neural-Based Metrics
- **COMET (Rei et al., 2020):** COMET (Crosslingual Optimized Metric for Evaluation of Translation) is a neural framework for MT evaluation that uses a multilingual pre-trained language model to score the MT output. It leverages deep representations to correlate closely with human judgments. However, it focuses primarily on semantic fidelity at segment-level granularity and does not inherently address style, context, or information completeness.

### 2.3 Multidimensional Frameworks
- **MQM (Multidimensional Quality Metrics):** MQM provides a comprehensive framework for defining and categorizing translation quality issues. It allows for a detailed analysis of different error types and their severity.
- **DQF (Dynamic Quality Framework):** DQF, developed by TAUS, is another framework for translation quality evaluation that integrates with translation workflows. It allows users to customize quality criteria based on content type, audience, and purpose.

However, both MQM and DQF typically require manual annotation, limiting scalability and not systematically addressing hallucinations, omissions, or contextual dependencies. In essence, no existing metric fully addresses the complexity of modern MT outputs. CATER draws inspiration from these approaches but transcends their constraints by employing an LLM-based, prompt-driven evaluation method. This approach enables comprehensive, scalable, and automated quality assessment.

# 3. The CATER Framework

CATER advances beyond single-reference and one-dimensional constraints by evaluating translation quality across five key dimensions. Each dimension resonates with MQM categories and dovetails with DQF's workflow-oriented perspective:

## 3.1 Linguistic Accuracy (LA)

Reflecting MQM's "Fluency" and "Grammaticality," LA assesses basic linguistic correctness—grammar, orthography, punctuation, and morphology. While seemingly fundamental, poor LA undermines trust and readability. Even minor linguistic errors damage credibility. Generally weighted moderately, but can be deprioritized if semantics or information fidelity outweigh stylistic nuances.

## 3.2 Semantic Accuracy (SA)

Aligned with MQM's "Accuracy," SA focuses on ensuring that the source text's meaning is faithfully conveyed. SA identifies mistranslations, semantic distortions, and hallucinations—cases where the translated text introduces meanings not present in the original or alters the intended sense. Unlike IC, which specifically addresses the omission or addition of critical factual information, SA targets issues where the semantic relationships or conceptual fidelity are compromised without necessarily losing or altering key facts. In other words, SA deals with incorrect or skewed interpretations of the source's meaning rather than missing or extra factual elements. This makes SA critical in domains where capturing subtle nuances and avoiding semantic shifts is paramount. Often heavily weighted in high-stakes contexts, any semantic distortion can undermine the translation's purpose and clarity.

## 3.3 Contextual Fit (CF)

Expanding upon MQM's discourse-level coherence, CF ensures that the translation's internal logic, pronoun reference, temporal consistency, and narrative integrity are preserved. For long-form texts or domain-specific documentation, CF becomes indispensable. May be weighted highly in contexts where narrative integrity or logical consistency is imperative.

## 3.4 Stylistic Appropriateness (STA)

Related to MQM's "Style" and DQF's adaptation mechanisms, STA ensures the tone, register, and style match the intended audience and purpose. Appropriate style can be critical in branding, literary translation, or culturally sensitive communication. Critical in texts where tone and audience appropriateness define success—literary works, diplomatic communications, or branding materials.

## 3.5 Information Completeness (IC)

Complementing MQM's focus on omissions and additions, IC addresses whether all critical facts,

figures, and domain-specific elements remain intact. Especially vital for legal, medical, technical, or safety-related documents, IC ensures that the translation neither omits nor distorts essential information. Indispensable in domains where losing or altering facts can have severe consequences. Frequently assigned high Weight in legal, medical, or technical fields.

**3.6 Weighting and Customization**

CATER allows for domain- and project-specific weighting. A technical manual might prioritize IC and SA, while a marketing text might emphasize STA. This flexibility transforms CATER from a fixed metric into a versatile evaluation platform aligning with diverse user priorities.

**3.7 LLM-Powered Prompt-based Evaluation**

CATER's key innovation lies in its operationalization: the evaluation is performed by prompting an LLM with a structured set of instructions (see Appendix). Given the source and translated texts, the LLM identifies "Words to Correct" for each category, computes the Edit Ratio (ER%), and converts these values into Category Scores and an Overall Score. By relying on the LLM's inherent language understanding, CATER not only bypasses the need for reference translations but also exploits the LLM's adaptability. With minor prompt edits, the framework can shift weighting, emphasize particular dimensions, or focus on a specific domain.

## 4. Scoring Methodology: ER%, Category Scores, and Overall Score

Each dimension yields a "Words to Correct" count, signifying the minimal textual edits needed to meet acceptable standards. From these counts, we derive the Category Edit Ratio (ER%):

$$ER\%_{category} = (Words\ to\ Correct_{category} / Original\ Word\ Count) * 100$$

A small ER% indicates fewer necessary edits, implying higher quality in that dimension. For interpretability, ER% values are mapped onto a 0–100 Category Score, with 100 indicating near-perfection and lower scores indicating greater issues. Adjusting each category's Weight modifies the impact of its ER% on the Category Score, ensuring that dimensions vital to the user's objectives carry greater influence on the final evaluation.

Summing weighted Category Scores produces an Overall Score (0–100) and aggregating ER%s yields an Overall ER%. While the Overall Score provides a top-level quality indicator, Category Scores and ER%s supply the granularity needed for targeted improvements. This hierarchical approach—from dimension-specific diagnostics to holistic assessments—facilitates both high-level comparison and in-depth quality management.

# 5. Advantages of an LLM-Driven Approach

### 5.1 Immediate Global Applicability

By harnessing an LLM's pre-trained multilingual capabilities, CATER is immediately applicable to virtually any language pair. Users need only supply the source and target texts plus a suitable prompt. This universality is a departure from traditional metrics requiring language-specific resources.

### 5.2 Flexible Customization and Scalability

CATER's prompt-based design allows evaluators to adjust weights per category, incorporate domain-specific guidelines, or add new quality dimensions without retraining models or collecting parallel data. The scalability and flexibility of this prompt-driven approach make CATER particularly appealing to large-scale translation agencies, global enterprises, and research institutions that handle a wide range of content types and linguistic profiles.

### 5.3 Real-time Iteration and Refinement

When evaluation outputs seem off or misaligned with user priorities, a simple prompt modification can produce a revised scoring scheme. This iterative cycle encourages a more interactive relationship between the evaluator and the metric, bridging conceptual frameworks and practical translation workflows.

### 5.4 Eliminating Reference Dependency

Unlike BLEU or TER, which depend on carefully constructed reference translations, CATER uses the LLM's contextual understanding to measure quality directly from the source and target texts. This enables evaluation of creative or user-tailored translations, as well as detection of subtle errors that might not appear if relying on a single "correct" reference.

# 6. Example Application and Illustrative Scenario

## 6.1 Example Application and Illustrative Scenario

Consider a scenario where a source text is a political speech intended to inspire trust, mobilize voters, and evoke a sense of collective purpose. The original passage envisions a brighter future on Inauguration Day, emphasizes scientific integrity, environmental stewardship, economic uplift, and social equality, culminating in a call-to-action to vote decisively and leave no doubt about collective identity and values. The translated text, however, may be grammatically correct yet fail to fully capture the intended meaning, style, context, and completeness.

## 6.2 CATER in Action (Political Speech Example):

**Original Text (Excerpt):**

"Imagine January 20th, when we swear in a President and Vice President who have a plan to get us out of this mess; who believe in science and have a plan to protect this planet for our kids; who care about working Americans and have a plan to help you start getting ahead; who believe in racial equality and are willing to do the work to bring us closer to an America where no matter what we look like, where we come from, who we love, or how much money we've got, we can make it if we try. All of that is possible. All of that is within our reach. If we pour all our effort into these last five days and vote up and down the ticket like never before, then we will elect Joe Biden and Kamala Harris. And we will leave no doubt about who we are and what this country stands for. Let's go."

**Translated Text (Excerpt):**

「1月20日の日を想像してください。この混乱から私たちを救い出す計画をもつ人物が大統領と副大統領として宣誓することを。科学を信じ、私たちの子供たちのためにこの星を守る計画を持つ人物。働くアメリカ人のことを気にかけ、人々が前進し始めることを助ける人物。人種の平等を信じ、見た目も、出身地も、誰を愛しているかも、どのくらい金を持っているかも問題にならない場所「アメリカ」に我々を導くため働きたいと願っている人物。我々が投票しさえすれば、この大統領を選べるのです。...がんばりましょう！」

**CATER Evaluation Summary:**

In applying CATER to this text, the evaluator identifies multiple issues across Semantic Accuracy (SA), Contextual Fit (CF), Stylistic Appropriateness (STA), and Information Completeness (IC). For instance:

**Semantic Accuracy (SA):**

The phrase "vote up and down the ticket like never before" is mistranslated as voting separately for various offices rather than emphasizing unified, comprehensive voting. This distorts the intended meaning. Another semantic issue is the portrayal of America as merely a "place" rather than conveying the inclusive societal vision the original text intends. The need to preserve "Let's go" as a forward-moving call-to-action rather than a generic "Let's do our best" further underscores semantic alignment challenges.

**Contextual Fit (CF):**
While the translation remains largely coherent at a sentence level, subtle shifts in phrasing turn an active, empowering assurance ("we will leave no doubt") into a passive statement of necessity ("no longer need to doubt"). This minor contextual misalignment reduces the text's forward momentum and confidence.

**Stylistic Appropriateness (STA):**
The speech aims to be inspirational and inclusive. Yet, certain choices in the translation (e.g., using "人物" for leaders) feel distant and impersonal. Adjusting terms to reflect personal engagement and urgency (e.g., replacing "がんばりましょう！" with "さあ、行きましょう！") restores the rallying tone. Removing unnecessary quotation marks around "アメリカ" maintains the original's inclusive, direct style.

**Information Completeness (IC):**
Critical clauses like "who are willing to do the work" and "we can make it if we try" are omitted, reducing the motivational and action-oriented quality. Also, adding conditions not present in the original ("我々が投票しさえすれば") introduces unauthorized constraints, altering the original message's intention and completeness.

After identifying errors and calculating Words to Correct for each dimension, the evaluator computes Edit Ratios (ER%) and Category Scores. For example, if SA requires several words to be changed or added, this could result in an SA ER% around 6.0% and a Category Score of 76/100. Similarly, STA and IC corrections yield their respective ER% and Score values. Summing up all the category-level results, the Overall Score might be as low as 32/100 with an Overall ER% of 17.6%, rating the translation as "Unusable."

This detailed, data-driven feedback pinpoints precisely where and how to improve the translation—adjusting certain phrases for better semantic fidelity, reinforcing stylistic resonance with the original's audience and tone, maintaining contextual logic, and ensuring all critical information is conveyed. By following CATER's recommendations, future revisions can significantly enhance the translation's communicative effectiveness and impact.

**6.3 CATER in Action (Literary Text Example):**

To illustrate CATER's adaptability, consider a literary text describing a winter landscape and nuanced character actions:

**Original Text (Excerpt) (from a famous Japanese novel "Yukiguni" by Yasunari Kawabata):**

「国境の長いトンネルを抜けると雪国であった。夜の底が白くなった。信号所に汽車が止まった。… 明りをさげてゆっくり雪を踏んで来た男は、襟巻で鼻の上まで包み、耳に帽子の毛皮を垂れていた。」

**Translated Text (Excerpt) (translated by Edward G. Seidensticker):**

"The train came out of the long tunnel into the snow country. The earth lay white under the night sky. The train pulled up at a signal stop. A girl who had been sitting on the other side of the car came over and opened the window in front of Shimamura..."

**CATER Evaluation Summary:**

In this case, the translation omits rich descriptive passages, lacks the subtle atmospheric tone, and fails to convey crucial narrative elements:

**Semantic Accuracy (SA):**

The translation simplifies or omits nuanced actions, such as the girl leaning out and calling for the stationmaster. It also neglects the description of the cold air rushing in.

**Stylistic Appropriateness (STA):**

A literary text demands careful maintenance of atmosphere, rhythm, and sensory detail. The direct, prosaic phrasing ("The train came out…") loses the quiet, immersive quality of the original.

**Information Completeness (IC):**

Key descriptive details about the man stepping onto the snow, muffled in a scarf, are entirely missing. This omission deprives the reader of the atmospheric richness and contextual clues central to the literary tone.

After analysis, suppose the CATER calculations reveal extremely high ER% values for SA and IC due to massive omissions and stylistic mismatches. This might yield a Semantic Accuracy Score close to 0/100 and an IC score of 0/100. Combined, these negative results drive the Overall Score down to 0/100, classifying the translation as "Unusable."

Such a stark evaluation highlights critical shortcomings: the need to restore lost imagery, detail, and narrative coherence. Armed with these insights, translators or editors can reintroduce omitted elements, refine their lexical choices to honor the source text's literary style, and restore the subtle, layered meaning intended by the original author.

**6.4 Conclusion of Example Scenarios:**

These illustrative examples show how CATER moves beyond single-reference metrics and simplistic matching. By dissecting texts across five dimensions (LA, SA, CF, STA, IC) and calculating standardized metrics like ER% and Category Scores, CATER provides a structured, quantitative map of translation quality. It identifies where and how a translation fails to meet its communicative goals—be they political inspiration, literary atmosphere, or factual completeness—and offers a foundation for targeted improvements. In doing so, CATER aligns closely with MQM and DQF principles, extending their conceptual frameworks into a scalable, automated, and domain-adaptive evaluation system suitable for both research and professional practice.

## 7. Benchmarking and Future Research

CATER's conceptual architecture supports rigorous comparison with established metrics on standard datasets, including the WMT metrics shared tasks. While this paper focuses on conceptual foundations and illustrative scenarios rather than large-scale empirical validation, future work will systematically benchmark CATER against widely used metrics—such as BLEU, TER, and COMET—and measure its correlations with human judgments. Evaluators may apply CATER to standard evaluation sets from WMT (Barrault et al., 2020; Bojar et al., 2016) or test it across diverse domains and language pairs. As LLM architectures advance, the framework's sensitivity and discriminative power will likely improve, offering evaluations that become more consistent, contextually aware, and aligned with human expectations.

Industry practitioners can also leverage CATER for large-scale analyses of parallel corpora, using its multidimensional insights to refine their MT workflows, inform retraining strategies, or guide targeted improvements in system outputs. Beyond benchmarking, future research will explore prompt optimization techniques, domain-specific extensions, and seamless integration with translation pipelines. For example, enterprises could incorporate CATER into QA processes, using LLMs to detect errors in real-time, dynamically adjust quality thresholds, or ensure stylistic consistency that aligns with brand identities.

Crucially, CATER is openly released, and we invite practitioners, researchers, and language professionals worldwide to experiment with the framework. Their feedback and data-driven insights will inform ongoing refinements, fostering a global, collaborative evolution of CATER. Over time, such community-driven engagement will not only enhance CATER's utility but also advance our collective understanding of cross-linguistic, cross-cultural translation phenomena, ultimately contributing to more nuanced and effective machine translation evaluation practices.

## 8. Conclusion

CATER stands as a pioneering, LLM-driven, and multidimensional MT evaluation framework that unites the theoretical richness of MQM and DQF with TER's core concept of edit effort. By integrating these insights into a comprehensive, reference-independent model, CATER not only quantifies editorial needs across Linguistic Accuracy (LA), Semantic Accuracy (SA), Contextual Fit (CF), Stylistic Appropriateness (STA), and Information Completeness (IC), but also highlights improvement pathways and encourages evidence-based prioritization. This holistic approach embraces a breadth of translation phenomena that conventional single-reference metrics often overlook.

By leveraging prompt-based interaction with powerful LLMs, CATER seamlessly evaluates linguistic, semantic, contextual, stylistic, and informational dimensions without the need for reference translations or custom models. Its adaptability, scalability, and cross-linguistic applicability offer unparalleled advantages to translation researchers, localization engineers, and professional translators worldwide. With open availability and an envisaged global feedback loop, CATER promises continuous refinement and evolution.

As the community of language professionals, MT researchers, and translation technology providers collaborates to refine this framework, CATER is poised to become a cornerstone in translation studies and MT evaluation practice. By bridging conceptual rigor, practical adaptability, and immediate deployability, CATER helps usher in a new era of flexible, context-rich, and user-centered translation quality assessment—one that is more holistic, context-aware, style-sensitive, and information-conscious.

## Acknowledgments

CATER was originally conceived and developed solely by Kurando Iida, who independently designed, conceptualized, and refined the framework proposed in this paper. The initial proposal and draft were prepared by Kurando Iida and subsequently enhanced with the assistance of ChatGPT o1 Pro-mode. Implementation was carried out by ErudAite CTO Kenjiro Mimura. CATER is made freely available at https://erudaite.ai/cater for an indefinite period, though free access may be discontinued without notice. Through this open release, we aim to gather extensive feedback from users across linguistic and cultural contexts. This collective input will guide large-scale empirical research that refines the framework and expands our understanding of translation quality across all languages.

# Appendix: Sample Prompt

*You are an evaluation assistant applying the CATER framework to assess a given translation's quality along five dimensions:*

1. *Linguistic Accuracy (LA)*
2. *Semantic Accuracy (SA)*
3. *Contextual Fit (CF)*
4. *Stylistic Appropriateness (STA)*
5. *Information Completeness (IC)*

***Your Inputs**:*

- *Original Text (source_text): [Paste the original source text here]*
- *Translated Text (translation_text): [Paste the translated text here]*

***Your Task**:*

1. *For each of the five categories (LA, SA, CF, STA, IC), identify all relevant errors if any.*
    - *For each error, provide:*
        - *Location (quote the problematic segment)*
        - *Error Explanation (why it's an error under that specific category)*
        - *Suggested Correction*
        - *Words to Correct (an estimate of how many words must be changed, added, or removed)*
    - *If no errors are found for a category, state "No errors detected."*
2. *Summarize each category:*
    - *Total Errors Detected*
    - *Words to Correct (sum of all corrections in that category)*
    - *Types of Errors (e.g., grammatical errors for LA, omissions for IC, etc.)*
3. *Calculate the Edit Ratio (ER%) for each category:*

$$\text{ER\%}_{\text{category}} = \frac{\text{Words to Correct}_{\text{category}}}{\text{Original Word Count}} \times 100$$

*Round to one decimal place.*

4. *Convert each category's ER% into a Category Score (0–100) using a simple formula example (for illustration, you may assume fixed weights as follows):*
    - *LA Weight = 1*
    - *SA Weight = 4*
    - *CF Weight = 3*
    - *STA Weight = 2*

- IC Weight = 5

For each category:

$$\text{Score}_{\text{category}} = \max\{0, \lfloor(1 - (\text{ER\%}/100 \times \text{Weight})) \times 100\rceil\}$$

If Words_to_Correct=0, then ER%=0.0%, Score=100.

If ER%>100%, Score=0.

*(Note: The above formula and weights are an example. You may use the provided weights or another consistent set.)*

5. Compute the Overall Score:

$$\text{Overall Score} = (\text{LA\_Score} + \text{SA\_Score} + \text{CF\_Score} + \text{STA\_Score} + \text{IC\_Score}) - 400$$

If the result is negative, set Overall Score=0.

If all categories are perfect (100 each), Overall Score=100.

6. Compute Overall ER% as the sum of each category's ER%.

**Final Output**:

- List all identified errors per category with details.
- Provide a summary (Words to Correct, ER%, Score) for each category.
- Show Overall Score and Overall ER%.

*Note*:

- This prompt is for demonstration and reference only.
- The output need not be stable or fully optimized.
- Actual implementations may involve more refined instructions and multiple prompts as shown in the main body of the paper.